\documentclass[10pt,twocolumn,letterpaper]{article}

\usepackage[pagenumbers]{cvpr} 

\usepackage{graphicx}
\usepackage{amsmath}
\usepackage{amssymb}
\usepackage{booktabs}

\usepackage[pagebackref,breaklinks,colorlinks]{hyperref}

\usepackage[capitalize]{cleveref}
\crefname{section}{Sec.}{Secs.}
\Crefname{section}{Section}{Sections}
\Crefname{table}{Table}{Tables}
\crefname{table}{Tab.}{Tabs.}
\usepackage[inline]{enumitem}

\def\cvprPaperID{*****} 
\def\confName{CVPR}
\def\confYear{2023}

\begin{document}

\title{ Emotional Reaction Intensity Estimation Based on Multimodal Data}

\author{Shangfei Wang, Jiaqiang Wu, Feiyi Zheng, Xin Li, Xuewei Li, Suwen Wang, Yi Wu, Yanan Chang, Xiangyu Miao\thanks{All authors have equal contribution.}\\
University of Science and Technology of China\\
}
\maketitle

\begin{abstract}
   This paper introduces our method for the Emotional Reaction Intensity (ERI) Estimation Challenge, in CVPR 2023: 5th Workshop and Competition on Affective Behavior Analysis in-the-wild (ABAW). Based on the multimodal data provided by the originazers, we extract acoustic and visual features with different pretrained models. The multimodal features are mixed together by Transformer Encoders with cross-modal attention mechnism. In this paper, 
   \begin{enumerate*}
   
   \item better features are extracted with the SOTA pretrained models. 
   \item Compared with the baseline, we improve the Pearson's Correlations Coefficient a lot. 
   \item We process the data with some special skills to enhance  performance ability of our model.
   \end{enumerate*} 

\end{abstract}

\section{Introduction}
\label{sec:intro}
Emotional reaction intensity estimation is an important part of recognizing and understanding people's feelings, especially for those robots that can interact with human beings. Thus, the interaction should not be dependent on the respective context, nor the human's age, sex, ethnicity, educational level, profession, or social position. And there are numerous and different dimensions of data during the interaction between machines and humans. In this paper, we present out methods for the Emotional reaction intensity estimation Challenge, in CVPR 2023: 5th Workshop and Competition on Affective Behavior Analysis in-the-wild (ABAW)\cite{0,1,2,3,4,5,6,7,8,9}. Our methods gets a good performance on the training, validation, and testing dataset.

The ERI\cite{10} estimation challenge, which targets at predicting the audience's intensities of seven different emotions. Each audience's reactions are aroused by the specific kind of videos. The reactions includes facial expressions, words, gestures. These multimodal data are usually disturbed by the
environment, such as light, noise, the quality of camera and so on. Even the different parts of the data can affect each other, so as to make the estimation more difficult. The ERI estimation needs us to utilize and analysis these multimodal data carefully to create a model, and the model should recognize human emotion states robustly by analyzing the audio information, visual information and gesture information. 

Our main contributions are as follows:
\begin{enumerate}
    \item We extract better audio features and video features with some pretrained models.
    \item In this challenge, compared with the baseline provided by the organizers, we improve the Pearson's Correlations Coefficient a lot.
    \item We process the data efficiently with some special fine tuning skills to enhance  performance ability of our model.
\end{enumerate}

The remaining paper is organized as follows. We introduce our pretrained models to extract audio and visual features in Section 2. And Section 3 describes our methods and framework. Then Section 4 presents the results based on the method before-mentioned. The conclusion is in Section 5. 

\section{Feature Extraction}
In this section we present our approach to extract visual and audio information. Mainly, we introduce the models used to extract features.

\subsection{Visual Information}

At first, we get the video frames from the video at intervals of 0.2 seconds. After getting the video frames, use DeepFace\cite{taigman2014deepface} to get the faces in the video frames in chronological order and save 224x224 face images. DeepFace is an efficient face recognition technology based on convolutional neural network proposed by Facebook AI, which can achieve high accuracy in different angles, lighting, and expressions. It is an essential technological advancement in the field of face recognition. 

After converting the video into a series of face images, we try out the model to obtain the visual features.

\textbf{FAb-Net}\cite{wiles2018self}: Facial Attributes-Net is trained to embed multiple frames from the same video face-track into a common low-dimensional space. The network learns a meaningful facial embedding that encodes information about the pose of the head, facial signs, and facial expressions without being supervised by any data from labels.

\begin{table}[!ht]
    \centering
    \caption{\textbf{MSE and PCC Based Models' Experiments Results on validation set}}
    \label{table 1}
    \begin{tabular}{lll}
    \hline
        Visual Feature & MSE & PCC \\ \hline
        FAb-Net & 0.2326 & 0.2543 \\ 
        DAN & 0.3581 & 0.3481 \\ 
        EfficientNet & 0.3591 & 0.3637 \\ 
        DAN+FAb-Net & 0.3718 & 0.3627 \\
        EfficientNet+FAb-Net & 0.3623 & 0.3701 \\ 
        EfficientNet+DAN & 0.3787 & 0.3826 \\ 
        FAb-Net+DAN+EfficientNet & \textbf{0.3894} & \textbf{0.3904} (0.3932) \\ \hline
    \end{tabular}
\end{table}

\begin{table}[!ht]
    \centering
    \caption{\textbf{Ensemble Model's Experiments Results on validation set }}
    \label{table 2}
    \begin{tabular}{ll}
    \hline
        Ensemble Model & PCC \\ \hline
        Train\_c43dd04b & 0.3904 \\ 
        + Train\_ed3eb7aa & 0.3979 \\ 
        + Train\_d2bdb6c2 & 0.4003 \\ 
        + Train\_1775502a & 0.4014 \\ 
        + Train\_ebf35676 & \textbf{0.4027} \\ \hline
    \end{tabular}
\end{table}

\textbf{EfficientNet}\cite{savchenko2021facial}: EfficientNet is a family of convolutional neural networks that achieve state-of-the-art accuracy and efficiency on various computer vision tasks. It was proposed by Google researchers in 2019 as a novel way of scaling CNNs based on a compound coefficient that balance the network depth, width, and resolution. The authors of EfficientNet use an automated neural architecture search(NAS) technique called MNAS to find the optimal base network architecture that can be scaled up efficiently. Similarly, we also use an automated fine tuning tools called Ray\cite{Ray} to help us find the most proper hyper parameters, which will be introduced in the detailed version of this paper.

\textbf{DAN}\cite{wen2021distract}: DAN leverages multi-head cross attention to capture subtle facial feature and reduce inter-class confusion. The network consists of three main components: a backbone network, a multi-head cross attention module, and a classifier. It concentrates on two problems: Firstly, multiple classed of facial expressions share inherently similar underlying facial appearance, and their differences could be subtle and localized. Secondly, existing FER methods tend to focus on the most salient regions of the face, such as the eyes and mouth, but neglect other regions that may contain useful information for FER.

Our experiments results on visual featured extracted with different pretrained models or tools are showing in Table \ref{table 1}. And the ensemble part of different models' experiments are showing in Table \ref{table 2}. 

\subsection{Audio Information}

According to audio information, we process the original wav files provided by organizers directly, with the help of several audio pretrained models or tools. In fact, we have tried a few different extracting methods to process audio data, while most of them perform not well. Therefore, we just pick some results that are not so bad, and the experiments results are showing in Table \ref{table 3}. Specifically, we design our own ResNet model just by replacing the conv2d to conv1d and using 7 residual blocks. Besides, we find that the modality of audio do not contribute to the final evaluation metric, while the single modality of video makes the proposed model perform better. It means that when acoustic features and visual features are fused together, the performance of our proposed model is not as good as before. 

\textbf{Wav2Vec2}: Wav2Vec2\cite{wav2vec2} is a large model pretrained and fine-tunued on 960 hours of Libri-Light and Librispeech\cite{kahn2020libri} on 16kHz sampled speech audio. Model was trained with Self-Training objective\cite{self}. Wav2Vec2.0 masks the speech input in the latent space and solves a contrastive task defined over a quantization of the latent representations which are jointly learned. We use the model provided by Facebook AI\cite{wav2vec2} to extract 768-dimensional acoustic features from the present audio files. 

\textbf{VGGish}: VGGish\cite{vggish} is a deep neural network model that was designed specifically for the task of audio classification. It was developed by researchers at Google and is based on the VGG network architecture. It has been used for a variety of audio classification tasks, including environmental sound classification and music classification. It has also been used to create embeddings for audio signals, which can be used for tasks such as audio similarity search and audio clustering. Given this, we also choose VGGish to extract 128-dimensional acoustic features.

\begin{table}[!ht]
    \centering
    \caption{\textbf{Experiments Results on validation set of Audio Modality}}
    \label{table 3}
    \begin{tabular}{lll}
    \hline
        Feature & model & PCC \\ \hline
        Wav2Vec2 & TE & 0.1048 \\ 
        Wav2Vec2 & ResNet & 0.0775 \\ 
        VGGish & TE & \textbf{0.1440} \\ 
        VGGish & ResNet & 0.08126 \\ \hline
    \end{tabular}
\end{table}

\section{METHODOLOGY}

\subsection{TE}
Our approach is mainly based on a CNN and a Transformer-Encoder\cite{9257201}.First, take the previously extracted features as input of the 1-dimensional temporal convolutional network to extract local temporal information. Then, the output of the 1-D CNN is fed into the transformer-encoder to achieve long-range dependencies with dynamic attention weights. Finally,  fully connected layers are used to inference the intensity of emotional reaction.

\subsection{Loss: MSE and PCC}

MSE can evaluate the degree of change in the data, and the smaller the value of the MSE, the better the accuracy of the predictive model in describing the experimental data. We use MSE to measure the matching degree between the predicted value and grand truth, so that the distribution between predicted value and grand truth can be similar.The Pearson correlation coefficient(PCC) is widely used to measure the degree of correlation between two variables, with values between -1 and 1.It is used to ensure the similarity of distribution between individual attributes.

\begin{figure}[t]
\centering
\includegraphics[width=.9\columnwidth]{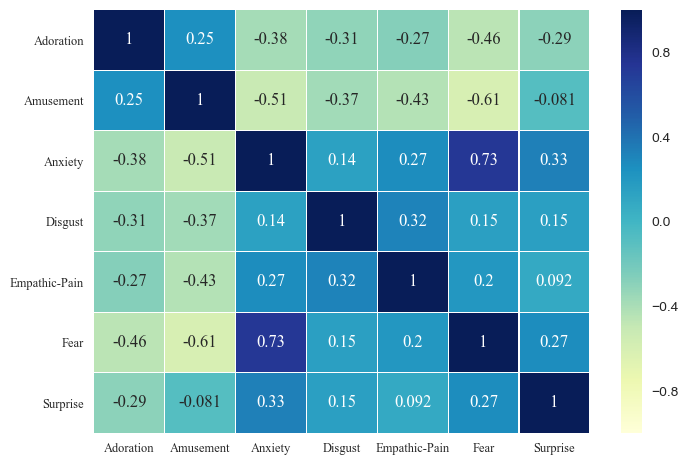} 
\caption{The correlation coefficient between labels in the training set.}
\label{fig:duration}
\end{figure}

\subsection{Training process and Integration strategy}
In the process of converting video frames to face images, we found that there were some videos that could not recognize faces, so we did some filtering of the training set and discarded the videos that could not recognize faces, but for the validation set, we scaled the video frames that could not recognize faces to the same size as the face images. The data in Table \ref{table 1} are obtained by training on the filtered training set and then validating on the full validation set, where the numbers in the last row in parentheses are the results obtained using the full training and validation sets.

Ray\footnote{https://www.ray.io/ray-tune} is a Python-based distributed computing framework that provides a set of tools and libraries for building and debugging distributed applications.  We use RAY to help us find better parameters to make the model work, where our settings in RAY are as follows: for the learning rate, we use it evenly between $1\times10^{-5}$ and $2\times10^{-4}$, for the batch size we set it between 8 and 32, and we limit the dimension  of the transformer hidden layer to 512 to 1024. We try 100 trials in RAY using MSE loss and PCC loss respectively, where each trial goes through a maximum of 10 epochs and how many epochs each trial goes through is determined by RAY, to obtain better hyperparameters through these trials.


\section{RESULTS}
We use the extracted face features as the visual information in the TE network framework and optimize them using MSE loss and PCC loss, respectively. As shown in Table \ref{table 1}, using only the visual information, we stitch the three visual features together and we achieve an average Pearson correlation coefficient of 0.3894 and 0.3904 in the validation set under MSE and PCC, respectively. For the audio information, after obtaining the audio features extracted by Wav2Vec2 and VGGish, we fed them into the TE and ResNet models, respectively, and optimized them using PCC losses. As shown in Table \ref{table 3}, using the TE network and the features extracted by VGGish, we achieve an average Pearson correlation coefficient of 0.1440 on the validation set.

\section{CONCLUSION}
In this paper, we introduce our method for the Emotional Reaction Intensity (ERI) Estimation Challenge, in CVPR 2023: 5th Workshop and Competition on Affective Behavior Analysis in-the-wild (ABAW). We utlize the acoustic features extracted by Wav2Vec2 and VGGish, and the visual features extracted by FAb-Net, EfficientNet and DAN. Based on the extracted features, we conduct experiments with TE model on each modality respectively. Besides, we train two models, which have the same network architecture, with different limitations. These two models have different performance on validation data, but when we ensemble these two models, then the ensemble model gets significant performance finally. The task's evaluation metric is Pearson's correlations coefficient, which is 0.4014 on validation set, outperforming the baseline a lot.
{\small
\bibliographystyle{ieee_fullname}
\bibliography{egbib}

\begin{thebibliography}{10}\itemsep=-1pt

\bibitem{wav2vec2}
Alexei Baevski, Yuhao Zhou, Abdelrahman Mohamed, and Michael Auli.
\newblock wav2vec 2.0: A framework for self-supervised learning of speech
  representations.
\newblock {\em Advances in neural information processing systems},
  33:12449--12460, 2020.

\bibitem{9257201}
Haifeng Chen, Dongmei Jiang, and Hichem Sahli.
\newblock Transformer encoder with multi-modal multi-head attention for
  continuous affect recognition.
\newblock {\em IEEE Transactions on Multimedia}, 23:4171--4183, 2021.

\bibitem{vggish}
Google.
\newblock Vggish, 2017.
\newblock
  \url{https://github.com/tensorflow/models/tree/master/research/audioset/vggish}.

\bibitem{kahn2020libri}
Jacob Kahn, Morgane Riviere, Weiyi Zheng, Evgeny Kharitonov, Qiantong Xu,
  Pierre-Emmanuel Mazar{\'e}, Julien Karadayi, Vitaliy Liptchinsky, Ronan
  Collobert, Christian Fuegen, et~al.
\newblock Libri-light: A benchmark for asr with limited or no supervision.
\newblock In {\em ICASSP 2020-2020 IEEE International Conference on Acoustics,
  Speech and Signal Processing (ICASSP)}, pages 7669--7673. IEEE, 2020.

\bibitem{0}
Dimitrios Kollias.
\newblock Abaw: Learning from synthetic data \& multi-task learning challenges.
\newblock {\em arXiv preprint arXiv:2207.01138}, 2022.

\bibitem{1}
Dimitrios Kollias.
\newblock Abaw: Valence-arousal estimation, expression recognition, action unit
  detection \& multi-task learning challenges.
\newblock In {\em Proceedings of the IEEE/CVF Conference on Computer Vision and
  Pattern Recognition}, pages 2328--2336, 2022.

\bibitem{5}
D Kollias, A Schulc, E Hajiyev, and S Zafeiriou.
\newblock Analysing affective behavior in the first abaw 2020 competition.
\newblock In {\em 2020 15th IEEE International Conference on Automatic Face and
  Gesture Recognition (FG 2020)(FG)}, pages 794--800.

\bibitem{7}
Dimitrios Kollias, Viktoriia Sharmanska, and Stefanos Zafeiriou.
\newblock Face behavior a la carte: Expressions, affect and action units in a
  single network.
\newblock {\em arXiv preprint arXiv:1910.11111}, 2019.

\bibitem{2}
Dimitrios Kollias, Viktoriia Sharmanska, and Stefanos Zafeiriou.
\newblock Distribution matching for heterogeneous multi-task learning: a
  large-scale face study.
\newblock {\em arXiv preprint arXiv:2105.03790}, 2021.

\bibitem{10}
Dimitrios Kollias, Panagiotis Tzirakis, Alice Baird, Alan~S. Cowen, and
  Stefanos Zafeiriou.
\newblock Abaw: Valence-arousal estimation, expression recognition, action unit
  detection \& emotional reaction intensity estimation challenges.
\newblock {\em ArXiv}, abs/2303.01498, 2023.

\bibitem{8}
Dimitrios Kollias, Panagiotis Tzirakis, Mihalis~A Nicolaou, Athanasios
  Papaioannou, Guoying Zhao, Bj{\"o}rn Schuller, Irene Kotsia, and Stefanos
  Zafeiriou.
\newblock Deep affect prediction in-the-wild: Aff-wild database and challenge,
  deep architectures, and beyond.
\newblock {\em International Journal of Computer Vision}, pages 1--23, 2019.

\bibitem{6}
Dimitrios Kollias and Stefanos Zafeiriou.
\newblock Expression, affect, action unit recognition: Aff-wild2, multi-task
  learning and arcface.
\newblock {\em arXiv preprint arXiv:1910.04855}, 2019.

\bibitem{4}
Dimitrios Kollias and Stefanos Zafeiriou.
\newblock Affect analysis in-the-wild: Valence-arousal, expressions, action
  units and a unified framework.
\newblock {\em arXiv preprint arXiv:2103.15792}, 2021.

\bibitem{3}
Dimitrios Kollias and Stefanos Zafeiriou.
\newblock Analysing affective behavior in the second abaw2 competition.
\newblock In {\em Proceedings of the IEEE/CVF International Conference on
  Computer Vision}, pages 3652--3660, 2021.

\bibitem{Ray}
Philipp Moritz, Robert Nishihara, Stephanie Wang, Alexey Tumanov, Richard Liaw,
  Eric Liang, William Paul, Michael~I. Jordan, and Ion Stoica.
\newblock Ray: {A} distributed framework for emerging {AI} applications.
\newblock {\em CoRR}, abs/1712.05889, 2017.

\bibitem{savchenko2021facial}
Andrey~V Savchenko.
\newblock Facial expression and attributes recognition based on multi-task
  learning of lightweight neural networks.
\newblock In {\em 2021 IEEE 19th International Symposium on Intelligent Systems
  and Informatics (SISY)}, pages 119--124. IEEE, 2021.

\bibitem{taigman2014deepface}
Yaniv Taigman, Ming Yang, Marc'Aurelio Ranzato, and Lior Wolf.
\newblock Deepface: Closing the gap to human-level performance in face
  verification.
\newblock In {\em Proceedings of the IEEE conference on computer vision and
  pattern recognition}, pages 1701--1708, 2014.

\bibitem{wen2021distract}
Zhengyao Wen, Wenzhong Lin, Tao Wang, and Ge Xu.
\newblock Distract your attention: Multi-head cross attention network for
  facial expression recognition.
\newblock {\em arXiv preprint arXiv:2109.07270}, 2021.

\bibitem{wiles2018self}
Olivia Wiles, A Koepke, and Andrew Zisserman.
\newblock Self-supervised learning of a facial attribute embedding from video.
\newblock {\em arXiv preprint arXiv:1808.06882}, 2018.

\bibitem{self}
Qiantong Xu, Alexei Baevski, Tatiana Likhomanenko, Paden Tomasello, Alexis
  Conneau, Ronan Collobert, Gabriel Synnaeve, and Michael Auli.
\newblock Self-training and pre-training are complementary for speech
  recognition.
\newblock In {\em ICASSP 2021-2021 IEEE International Conference on Acoustics,
  Speech and Signal Processing (ICASSP)}, pages 3030--3034. IEEE, 2021.

\bibitem{9}
Stefanos Zafeiriou, Dimitrios Kollias, Mihalis~A Nicolaou, Athanasios
  Papaioannou, Guoying Zhao, and Irene Kotsia.
\newblock Aff-wild: Valence and arousal ‘in-the-wild’challenge.
\newblock In {\em Computer Vision and Pattern Recognition Workshops (CVPRW),
  2017 IEEE Conference on}, pages 1980--1987. IEEE, 2017.

\end{thebibliography}
}

\end{document}